\documentclass[journal,twoside,web]{ieeecolor}
\usepackage{jsen}
\usepackage{cite}
\usepackage{amsmath,amssymb,amsfonts,bm,amssymb}
\usepackage{algorithmic}
\usepackage{graphicx}
\usepackage{textcomp}
\usepackage{wrapfig}
\usepackage{booktabs}
\usepackage{subfigure}
\usepackage{multirow}
\usepackage[final]{changes} 
\definechangesauthor[name={first_change}, color=blue]{A1} 
\def\BibTeX{{\rm B\kern-.05em{\sc i\kern-.025em b}\kern-.08em
    T\kern-.1667em\lower.7ex\hbox{E}\kern-.125emX}}
\definecolor{abstractbg}{rgb}{0.89804,0.94510,0.83137}
\begin{document}
\title{EMA-VIO: Deep Visual-Inertial Odometry with External Memory Attention}
\author{Zheming Tu*, Changhao Chen*, Xianfei Pan$^{\dag}$, Ruochen Liu, Jiarui Cui, Jun Mao
\thanks{The authors are with the College of Intelligence Science and Technology, National University of Defense Technology, Changsha, 410073, China.}
\thanks{Zheming Tu* and Changhao Chen* are co-first authors. (Email: tzm\_nudt@163.com, changhao.chen66@outlook.com)}
\thanks{Xianfei Pan$^{\dag}$ is the corresponding author. (Email: afeipan@126.com)}
\thanks{This work was supported by National Natural Science Foundation of China (NFSC) under the Grant Number of 62103427, 62073331, 62103430.}
}

\IEEEtitleabstractindextext{%
\fcolorbox{abstractbg}{abstractbg}{%
\begin{minipage}{\textwidth}%
\begin{wrapfigure}[15]{r}{2.9in}%
\includegraphics[width=2.9in]{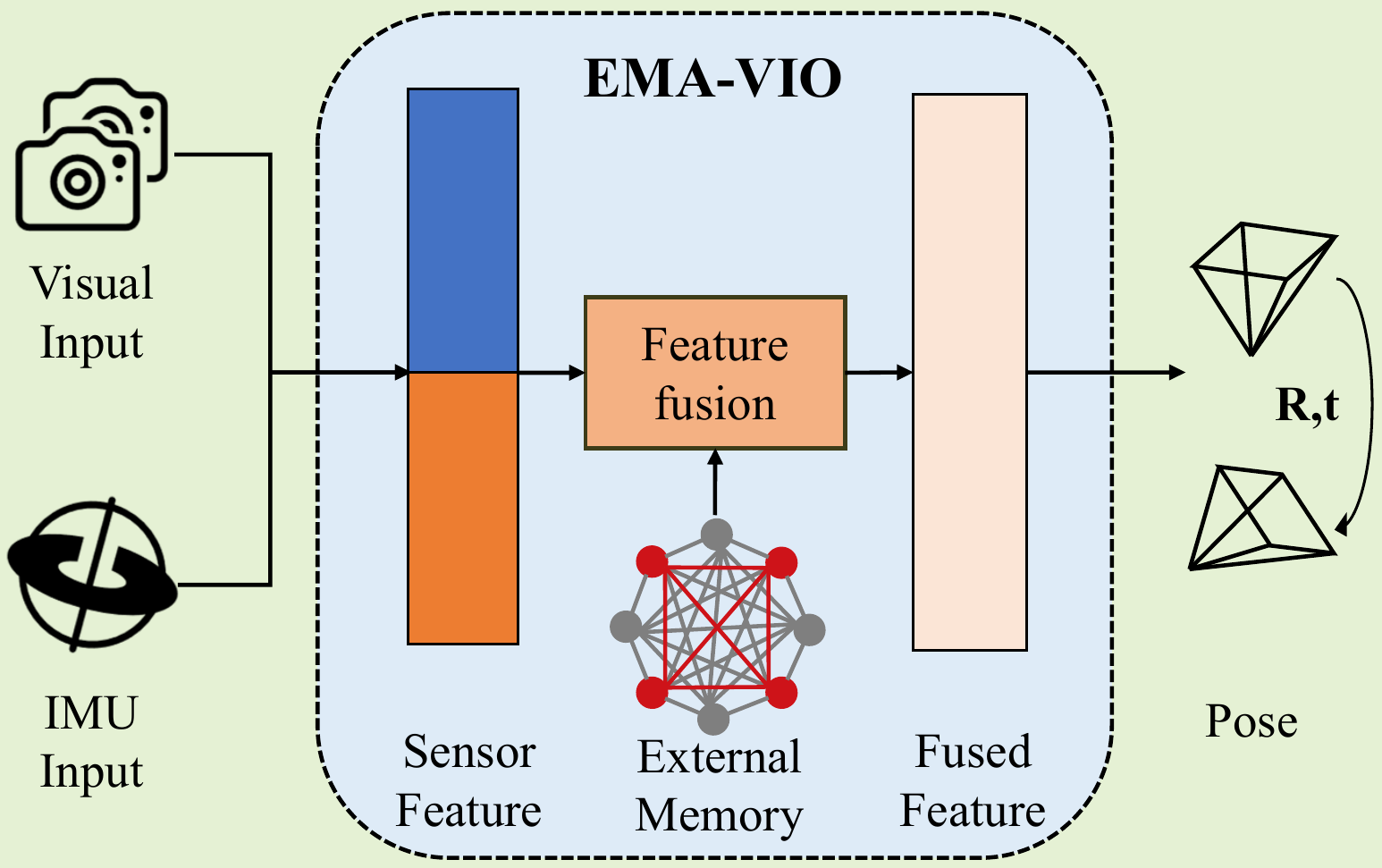}%
\end{wrapfigure}%
\begin{abstract}
Accurate and robust localization is a fundamental need for mobile agents. Visual-inertial odometry (VIO) algorithms exploit the information from camera and inertial sensors to estimate position and translation. Recent deep learning based VIO models attract attentions as they provide pose information in a data-driven way, without the need of designing hand-crafted algorithms. Existing learning based VIO models rely on recurrent models to fuse multimodal data and process sensor signal, which are hard to train and not efficient enough. We propose a novel learning based VIO framework with external memory attention that effectively and efficiently combines visual and inertial features for states estimation. Our proposed model is able to estimate pose accurately and robustly, even in challenging scenarios, e.g., on overcast days and water-filled ground , which are difficult for traditional VIO algorithms to extract visual features. Experiments validate that it outperforms both traditional and learning based VIO baselines in different scenes.
\end{abstract}

\end{minipage}}}

\maketitle

\section{Introduction}
\label{sec:introduction}
\IEEEPARstart{A}{ccurate} positioning plays a vital role in mobile agents to support many applications, e.g. self-driving vehicles, delivery robots, and augmented reality. It is difficult to achieve long-term localization only using a single sensor in complex environments. With a rich combination of sensors (e.g. camera, IMU, LIDAR) on mobile platforms, effective sensor fusion is always a central problem in robust designing localization and navigation system. 
However, the problem of sensor fusion is complicated due to real-world issues, such as the diversity of error characteristics , the spacial and temporal misalignment of different sensors. 
Due to the natural complementarity of camera and inertial measurement unit (IMU), researchers throw their interests on integrating camera and IMU as visual-inertial odometry (VIO) to provide pose information. Fully exploiting the properties and advantages of two sensors is important to improving the positioning accuracy and robustness. 

Existing VIO algorithms can be divided into traditional model based methods \cite{qin_vins-mono_2018,engel_direct_2017,campos_orb-slam3_2021}, and deep learning based methods \cite{clark_vinet_2017,yang_d3vo_2020}. Traditional model based VIOs rely on hand-designed feature detectors, and filtering \cite{Li2013b} or non-linear optimization \cite{qin_vins-mono_2018} based sensor fusion.
Learning based VIO models can extract features automatically, and learn pose from images and IMUs in an end-to-end data driven way \cite{clark_vinet_2017,chen_selectfusion_2019,Wei2020}. These methods have shown their superiority in robustly operating in challenging environments. 

However, previous learning based VIOs are not efficient enough and hard to train and infer. Effective features fusion is less explored, compared with much effort on feature learning. To solve the feature fusion problem more effectively and efficiently, we propose \textbf{EMA-VIO}, a novel framework that introduces \textbf{E}xternal \textbf{M}emory \textbf{A}ttention into learning based \textbf{VIO}s.
Previous normally leverage Long-short Term Memory (LSTM) network to fuse visual and inertial features \cite{clark_vinet_2017,chen_selectfusion_2019,Wei2020}. 
It is difficult for recurrent models to train and infer in parallel, as recurrent model has to process sequential data in order.
Recently, in natural language processing and image processing, there are growing interests in replacing recurrent model with attention mechanisms. Our proposed external memory attention based sensor fusion is able to improve the performance of VIOs, and reduces the memory and computation requirement. To further enhance the capacity of pose learning, we impose multistate pose constraints to refine the pose estimation. Experiments in real-world with our own robot platforms demonstrate that our EMA-VIO can operate in less-feature environments.

The main contributions of this paper can be summarized into the following three parts:
\begin{enumerate}
    \item we propose a novel external memory attention based fusion mechanism for VIO learning framework that effectively combines features from two sensors and improves the model performance.
    \item A multi-state geometric constraint is introduced into the pose loss function that speeds up training convergence and contributes more accurate pose results.
    \item Real-world experiments were conducted to evaluate our proposed learning based VIO model above our own robot platform. It shows that our learning model can perform well in challenging environments, i.e., overcast day and wet ground.
\end{enumerate}

\section{Related work}
\added{Recently, deep learning has been widely applied in solving real-world problems, e.g., COVID-19 diagnosis \cite{yao2022adad}.} Our work is mainly relevant to deep learning based visual positioning, deep learning based visual-inertial odometry and attention mechanisms. Thus, we discuss them in this section.
 
\subsection{Deep Learning based Visual Positioning} 
Learning based visual positioning methods have made great progress in recent years. DeepVO\cite{wang_deepvo_2017} proposes an end-to-end visual odometry based on deep neural networks, which uses Convolutional Neural Networks (ConvNets) to encode image sensor data into visual features, and a two-layer LSTM network to update visual features to provide pose in an end-to-end manner. There are some recent works that further improve the performance of learning model \cite{xue2019beyond,Zhan2018,undeepvo,zhan2020visual,KuoLLLCL20}.
Especially, \cite{brahmbhatt_geometry-aware_2018} propose a geometry-aware learning framework for camera localization that improves the accuracy capability of both visual odometry and relocalization. 
AtLoc \cite{wang_atloc_2020}, proposes a camera localization model guided by a self-attention mechanism, which introduces the encoded features into the attention mechanism to form attention matrix, and then the attention matrix is summed with the original visual features to form new features, which it improves the accuracy of localization. We introduce an external memory attention into learning based localization model which has never been seen in previous works.

\subsection{Deep Learning based Visual-inertial Odometry}
VINet\cite{clark_vinet_2017} is an end-to-end trainable visual-inertial odometry model, that takes sensor data from monocular camera and IMU as input, and adopts ConvNet and recurrent network to encode multi-modal features from visual and inertial information respectively. VINet concatenates the two kind of feature directly according to channel dimensions, and process features with recurrent model. 
To further study the multimodal fusion problem for VIO, \cite{chen2019selective} proposes a selective fusion mechanism for learning based VIO to fuse features from two sides. Compared with direct fusion, their proposed fusion method achieves better performance in localization and is more robust in dealing with multiple cases of sensor data degradations.
SelfVIO\cite{almalioglu_selfvio_2022} proposes a self-supervised trained monocular visual-inertial odometry for localization and depth prediction, 
but it can not provide pose in an absolute scale metric.
In \cite{liu_atvio_2021}, two convolutional neural networks are proposed for the feature extraction of inertial data, replacing the LSTM network in the inertial feature extraction part of selectfusion network \cite{chen2019selective}, and the extracted features are then spliced and interpolated to form the inertial feature matrix, after which the feature extraction is performed by the attention mechanism. However, all of these frameworks still rely recurrent models to fuse multimodal features and update system states, which are hard to train and limit the efficiency of entire framework.
\added{CodeVIO\cite{zuo2021codevio} proposes a lightweight, tightly-coupled visual-inertia odometry network along with depth network to provide accurate state estimates and dense depth estimates of the surroundings.}
\added{\cite{aslan2022visual} learns the pose of a drone in a more stable way, by denoising inertial data, adopting Inception-v3 to extract visual feature from two consecutive images, and producing pose via Gaussian Process Regression.}
\subsection{Attention mechanisms}
 There are growing interests in attention mechanisms in the deep learning community. Attention has been introduced into the encoder-decoder architecture to associate information in a long range.
\cite{mnih_recurrent_2014} propose to use attention mechanism into recurrent networks for image classification, which inspires the subsequent development of attention mechanisms. \cite{bahdanau_neural_2014} apply the attention mechanism to the natural language processing (NLP) domain, in which a sequence-to-sequence architecture with attention model is used for machine translation with astonishing results. Recently, the Transformer\cite{vaswani_attention_2017} model has attracted great attentions, in which authors use self-attention mechanisms to replace recurrent models in the neural language processing problems, showing superiority in translation quality while being more parallelizable and requiring significantly less time to train.
Self-attention has been used in visual processing \cite{wang2018non}. Furthermore, in order to solve the problem of excessive computation caused by the self-attention mechanism in visual tasks,
external attention mechanism is proposed by \cite{guo_beyond_2021} showing comparable or superior performance towards self-attention mechanisms with lower computational and memory costs.

\begin{figure*}[htbp!]
    \centering
    \includegraphics[scale=.6]{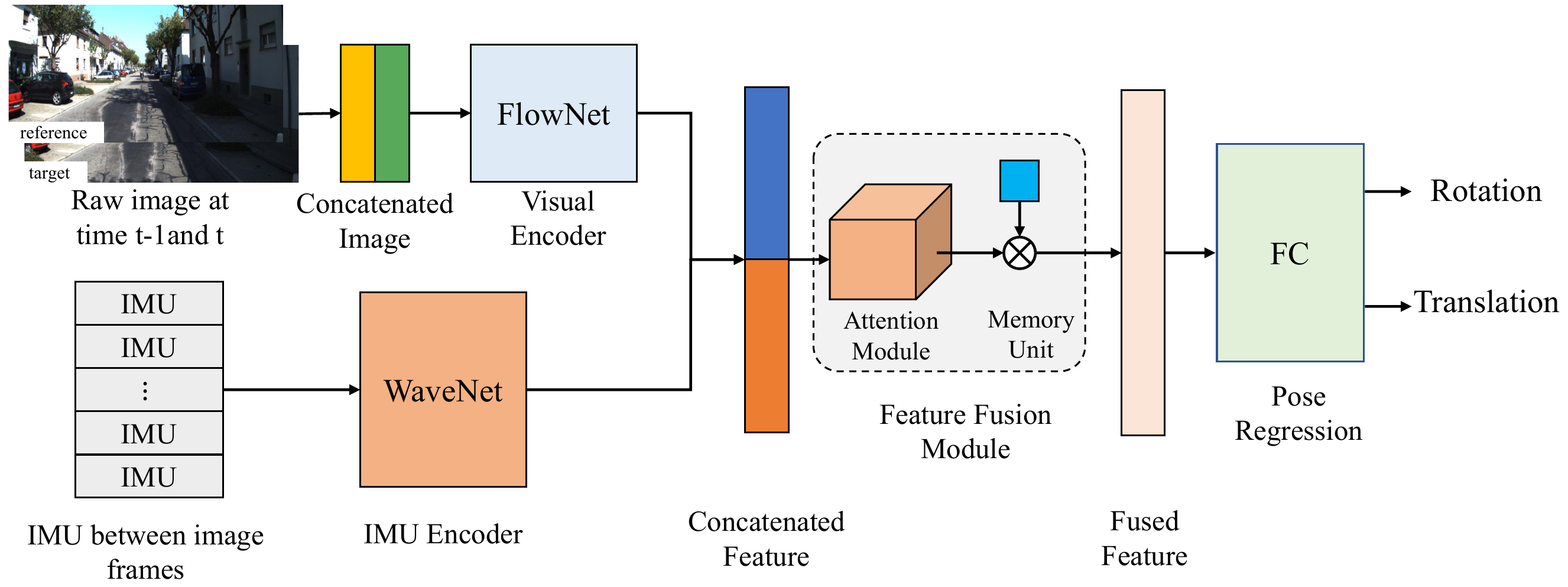}
    \caption{The architecture of our proposed deep visual-inertial odometry model.}
    \label{fig:architecture}
\end{figure*}

\section{Deep Visual-Inertial Odometry}
In this section, we discuss our proposed deep learning based VIO model in details. The framework of our approach is shown in Figure \ref{fig:architecture}. Our framework mainly includes feature encoders, external memory attention based feature fusion module, and pose regressor. The network inputs a pair of images and a sequence of IMU data, and outputs pose transformation.
To effectively fuse visual and inertial features, we introduce an attention mechanism with memory unit assistance , unlike the previous works commonly using LSTM \cite{clark_vinet_2017,chen_selectfusion_2019,Wei2020}. 
\subsection{Feature Encoders}
\subsubsection{Visual Feature Encoder}
In the visual feature encoder,
we aim to extract geometry-meaningful features from a combination of two image, so that FlowNet \cite{ilg_flownet_2017} is adopted to provide features relevant to optical flow and motion estimation.
There are six convolutional layers, and each convolutional layer is followed by a Relu activation layer. Two consecutive frames are concatenated channel-wise and input to the network. The first image can be defined as reference image frame $\mathbf{P}_{\text{ref}}$ and the second image as target image frame $\mathbf{P}_{\text{tar}}$. The front-end convolution part of FlowNetS is used to extract features from the combined images to form the feature representation, after which a pooling layer to downsample the high-dimensional feature representation and a fully connected layer maps it into a visual feature vector $\mathbf{F}_v$.   
\begin{equation}
    \mathbf{F}_V=\text{FlowNet}([\mathbf{P}_\text{ref}; \mathbf{P}_\text{tar}]).
\label{equ:visual}\end{equation}

\subsubsection{Inertial Feature Encoder}
\begin{figure}[htbp!]
    \centering
    \includegraphics[scale=.4]{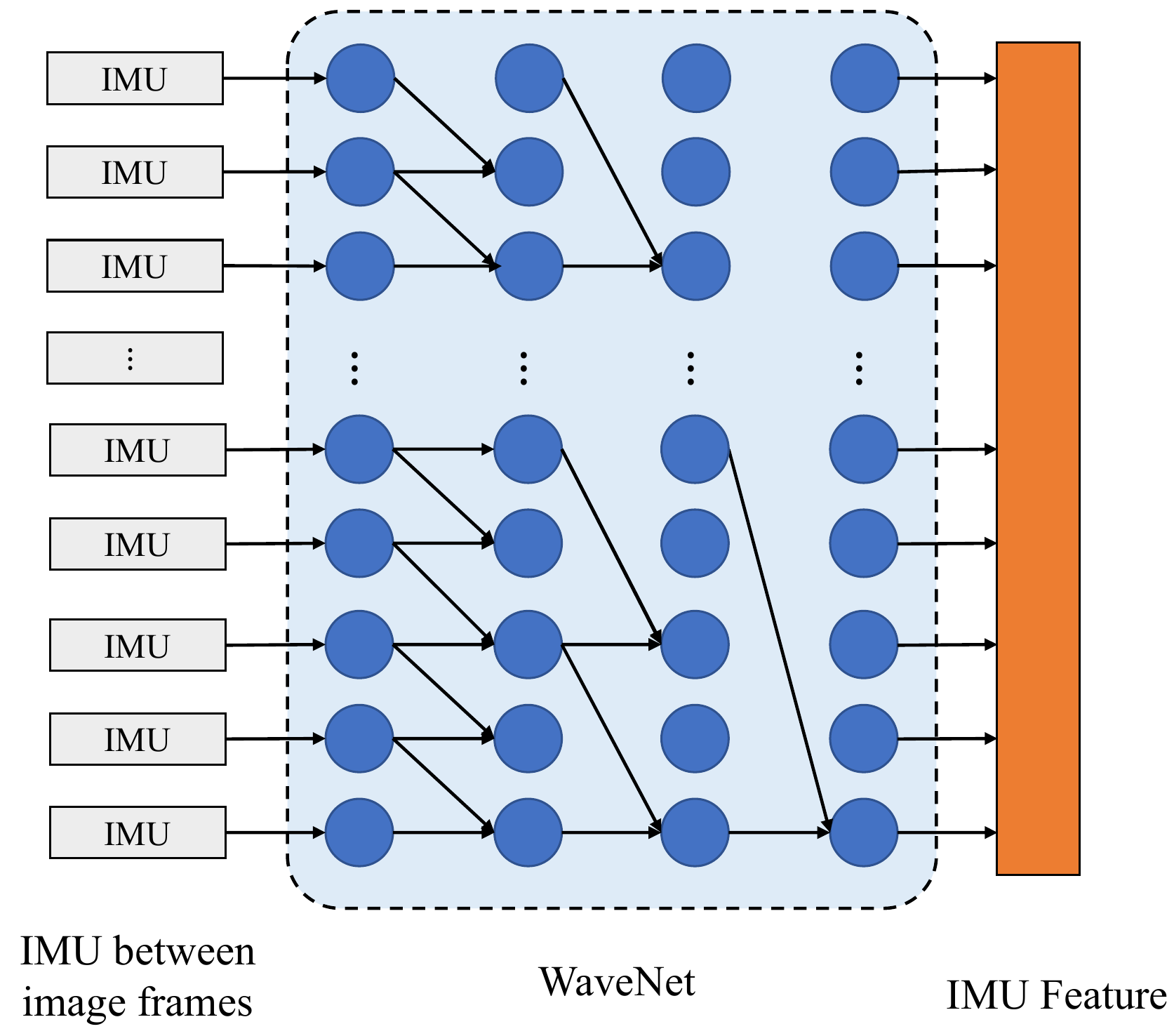}
    \caption{The inertia feature extraction block which is composed of our proposed model.}
    \label{fig:wavenet}
    \end{figure}

Instead of using LSTM to process inertial data, we adopt
a WaveNet style module to extract inertial features \cite{oord2016wavenet}.
Here, we construct an inertial feature encoder based on dilated causal convolutional neural networks. As shown in Figure \ref{fig:wavenet},
in the dilated causal convolutional network, the prediction of all time steps can be performed in parallel. 
And the training process of the network is faster and more efficient because the circular connection structure of recurrent model is eliminated. 
The inertial data between adjacent image frames are as input, we use causal convolutional network to process a sequence of inertial data to form inertial feature vectors. The entire convolutional network has four layers, and each convolutional layer is followed by a gate activation unit:
\begin{equation}\textbf{z}=\text{tanh}(\textbf{W}_{f,k}*\textbf{x})\odot\sigma(\textbf{W}_{g,k}*\textbf{x}).\label{activation}\end{equation}
where \added{$\textbf{x}$ represents the output of each causal convolutional layer,} $\textbf{W}$ denotes the weight matrix of the convolution kernel, \added{$g$ and $f$ represent gate and filter respectively, and $k$ is the layer index}, $*$ denotes the convolution operation, $\odot$ denotes the elemental multiplication operation, $\sigma(\cdot)$ denotes the elemental sigmoid activation function, and $\text{tanh}$ denotes the hyperbolic tangent activation function. \added{The inertial feature is obtained from the sum of the result of each layer.}

In practice, 
we use one-dimensional convolutional network with 64 kernels to convolve the inertial measurement data $\textbf{I}$ in a sequential order, perform activation and then enter feature representation into the next layer of convolutional network. The convolutional networks are connected by both residual and skip connections to speed up the convergence of the model and allow the gradient to be transmitted to deeper levels of the model. The convolutional kernel interval between each layer is expanded by a power of 2, which expands the perceptual field. Finally, we can obtain the inertia feature $\textbf{F}_I$. 

\begin{equation}
\textbf{F}_I=\text{WaveNet}(\textbf{I}).
\label{inertial}
\end{equation}

    \begin{figure*}[h!]
    \centering
    \includegraphics[scale=.6]{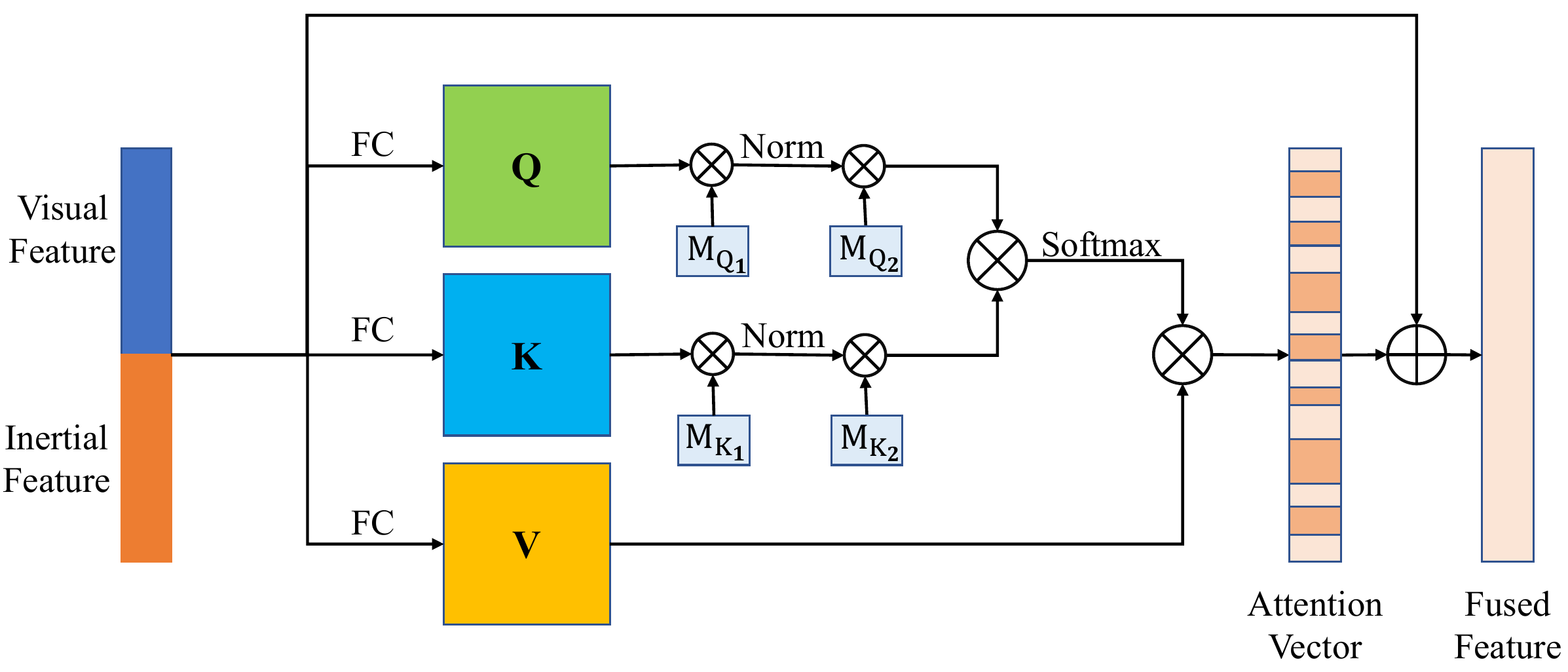}
    \caption{The attention block of our proposed model.}
    \label{fig:attention}
    \end{figure*}

\subsection{Memory Attention based Feature Fusion}
 We combine the feature vectors from two feature encoders in their channel dimension $[\textbf{F}_V; \textbf{F}_I]$, and then selectively fuse the feature vectors using attention mechanism to obtain the fused feature vector $\textbf{F}_{\text{att}}$.
 
\begin{equation}
    \textbf{F}_{\text{att}}=\text{Attention}([\textbf{F}_V; \textbf{F}_I]).
    \label{featurefusion}
\end{equation}

In recent years, attention mechanism becomes a hot topic in both natural language processing and computer vision. 
Attention mechanism can be used as a good solution to associate information. \textit{Our proposed memory attention based feature fusion is based on non-local self-attention with introduced memory variables to further enhance its capacity to fuse features.}

\subsubsection{Self Attention}
In the self-attentive mechanism, the input feature matrix $\mathbf{F}$ is converted to a query matrix $\mathbf{Q}$, a key matrix $\mathbf{K}$, and a value matrix $\mathbf{V}$ through a linear layer according to Eq. \ref{eq:QKV}:
\begin{equation}
\mathbf{Q}=\mathbf{W}_{q} \mathbf{F},\ \ \mathbf{K}=\mathbf{W}_{k} \mathbf{F},\ \ \mathbf{V}=\mathbf{W}_{v} \mathbf{F},
\label{eq:QKV}
\end{equation}
where $\mathbf{W}_{q}$, $\mathbf{W}_{k}$ and $\mathbf{W}_{v}$ are the transform matrix of linear layers.

Then we calculate the self-attention matrix $\mathbf{A}$ via equation \ref{eq:attention}, which represents the attention weights of the neural network at the corresponding positions of the $\mathbf{A}$ matrix.

\begin{equation}
\mathbf{A} = \text{Softmax}(\mathbf{Q} \mathbf{K}^T) \mathbf{V}.
\label{eq:attention}
\end{equation}

After that, residual connection is performed according to Eq. \ref{featurenew} to connect the attention matrix with the original feature matrix to obtain the feature matrix after reweighting by the attention mechanism, which can be further used for the subsequent positional decoding and pose regression.
\begin{equation}
\mathbf{F}_\text{att}=\alpha(\mathbf{A})+\mathbf{F}.
\label{featurenew}
\end{equation}
where the linear layer $\alpha(\mathbf{A})=\mathbf{W}_{\alpha} \mathbf{F}$ denotes a learnable scaled self-attention vector.

\subsubsection{External Memory Aided Self-attention}
Inspired by the external attention mechanism, we introduce the memory variables into the self-attention mechanism to form our attention based fusion module as shown in Fig. \ref{fig:attention}. 
Based on the self-attention mechanism, two memory units $\mathbf{M}_1$ and $\mathbf{M}_2$ are added to the channels of calculating the $\mathbf{Q}$ and $\mathbf{V}$ matrix of the self-attention.
These two external learnable memory variables can be implemented by two linear layers. Therefore, these two memory variable parameters can be updated during backpropagation.
\begin{equation}
    \mathbf{Q}_\text{att}=\text{Norm}(\mathbf{Q} {\mathbf{M}_{{Q}_1}^T}) \mathbf{M}_{{Q}_2}.
    \label{eq:Qnewattention}
\end{equation}
\begin{equation}
    \mathbf{K}_\text{att}=\text{Norm}(\mathbf{K} \mathbf{M}_{{K}_1}^T) \mathbf{M}_{{K}_2},
    \label{eq:Knewattention}
\end{equation}
where $\mathbf{M}_{{Q}_1}$, $\mathbf{M}_{{Q}_2}$, $\mathbf{M}_{{K}_1}$ and $\mathbf{M}_{{K}_2}$ are linear matrix to update $\mathbf{Q}$ and $\mathbf{K}$.
 Also, since these two variables are not determined by a particular sample, but are bounded by the samples in the whole dataset, they act as a regularizer for the samples in the whole dataset, which in turn improves the generative ability of the attention mechanism and the pose estimation accuracy.

The introduced memory units can be viewed similar to the hidden states of recurrent model, that store and maintain the history information of a sequence.
With our proposed fusion mechanism, we are able to replace the LSTM network commonly used in previous works. Our experiments show that this mechanism further improves the model performance and meanwhile reduces computational and memory requirement.

\subsection{Pose Regressor with Geometric Multi-state Constraint}

After feature fusion, a pose regressor is designed to map the fused features to relative pose vector between two frames, i.e., 3-dimensional position displacement $\mathbf{t}$ and 3-dimensional rotation $\bm{\psi}$.
It consists of two linear layers with a Relu activation function between them to perform pose regression,
\begin{equation}
    (\mathbf{t}, \bm{\psi})=f_\text{pose}(\mathbf{F}_\text{att}).
    \label{regression}
\end{equation}

\begin{figure}[htbp!]
    \centering
    \includegraphics[scale=.65]{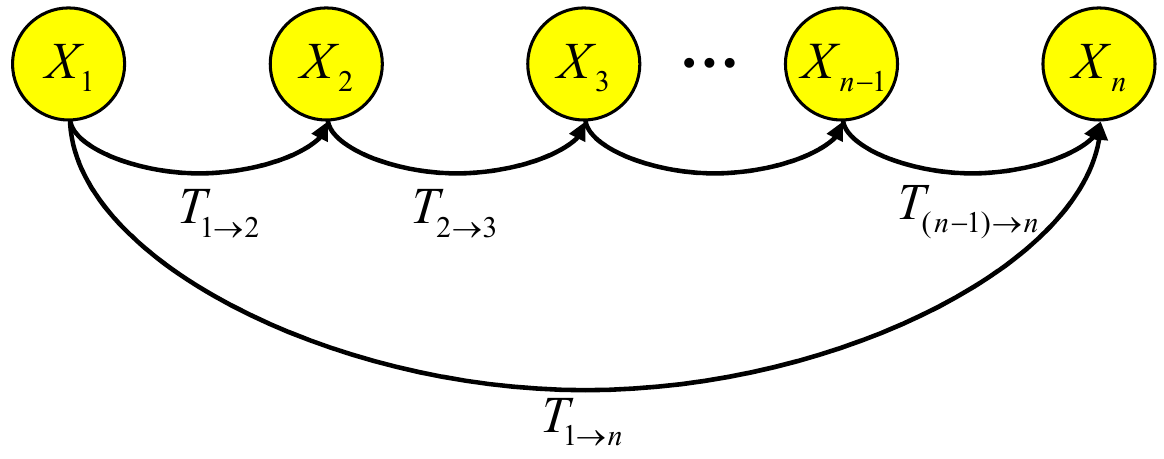}
    \caption{The geometric multi-state constraint of the sequence.}
    \label{fig:mc}
    \end{figure}
The geometric relation between poses is important information to pose update, as illustrated in Fig. \ref{fig:mc}. Thus, our pose regressor further exploits this geometric multi-state constraint into pose optimization.
The position displacement and rotation vectors $(\mathbf{t}, \bm{\psi})$ between two states are use to form transformation matrix $\mathbf{T}$ representing motion transformation,
\begin{equation}
    \mathbf{T} = \begin{bmatrix}
                    \mathbf{R} & \mathbf{t}\\
                    \mathbf{0} & \mathbf{1}\\
                 \end{bmatrix}
\end{equation}
where $\mathbf{R}$ is rotation matrix which can be derived from Euler angle based rotation vector $\bm{\psi}$.

In our model, the length of image sequence is chosen as $n$, and the pose transformation from the i-th frame to the j-th frame is expressed as $\mathbf{T}_{i \rightarrow j}$,$\mathbf{T} \in SE(3)$. 
The pose translation from the first frame to the last frame of the sequence can be expressed as $\mathbf{T}_{1 \rightarrow n}$. 
The relative poses between frames have to satisfy the following multi-state constraint:
\begin{equation}\mathbf{T}_{1 \rightarrow 2} \cdot \mathbf{T}_{2 \rightarrow 3} \cdots \mathbf{T}_{(n-1) \rightarrow n}=\mathbf{T}_{1 \rightarrow n}.\label{mulitistate}\end{equation}
This multi-state geometric constraint will be incorporated into loss function to refine pose estimation.

\subsection{Loss Function and Optimizer}
Finally, a loss function is defined to optimize the entire framework to estimate poses. Our loss function $L_\text{total}$ consists of a per-frame loss $L_\text{frame}$ and a sequence loss $L_\text{seq}$: 
\begin{equation}
L_\text{total}=L_\text{frame}+L_\text{seq}.
\label{loss}
\end{equation}
Based on this loss function, we adopt the ADAM optimizer to recover the optimal parameters of neural network framework.

\subsubsection{Per-frame loss}
The per-frame loss represents the loss of estimated poses and ground-truth poses between two frames.
Here we use the Euler angle $\bm{\psi}$ to denote the rotation between two image frames and the relative translations in metric units $\mathbf{t}$. 
And the root mean square error (RMSE) is adopted, 
\begin{equation}L_\text{frame}=\text{RMSE}(\mathbf{t}_\text{frame}-\hat{\mathbf{t}}_\text{frame})+\lambda_1 \cdot \text{RMSE}(\bm{\psi}_\text{frame}-\hat{\bm{\psi}}_\text{frame})\label{lossadj}\end{equation}
where $(\hat{\mathbf{t}}, \hat{\bm{\psi}})$ is the ground-truth pose, $\lambda_1$ is a scale-factor to balance the per-frame loss between translation and rotation.

\subsubsection{Sequence Loss}
The sequence loss exploits the multi-state geometric constraint to define a loss function between the ground truth and the predicted values of the whole sequence. 
Following the Equation \ref{mulitistate}, the sequence transformation matrix representing the relative motion from first frame to the last frame of a sequence can be acquired, and the translation and rotation $(\mathbf{t}_\text{seq}, \bm{\psi}_\text{seq})$ of a sequence can be derived correspondingly. The ground-truth of a sequence motion transformation denotes $(\hat{\mathbf{t}}_\text{seq}, \hat{\bm{\psi}}_\text{seq})$.
\begin{equation}
L_\text{seq}=\text{RMSE}(\mathbf{t}_\text{seq}-\hat{\mathbf{t}}_\text{seq})+\lambda_2 \cdot \text{RMSE}(\bm{\psi}_\text{seq}-\hat{\bm{\psi}}_\text{seq}).
\label{lossseq}
\end{equation}
where RMSE denotes the root mean square error function and $\lambda_2$ denotes the scale factor between the displacement error and the rotation error.

\section{Experiments}
In this section, we first evaluated our proposed deep VIO model on the KITTI dataset, which is public available, and commonly used as benchmark to evaluate both classical and learning based positioning algorithms. Then, we conducted experiment with our own robot platform to collect data and evaluate our proposed model in challenging environments.
The framework of our method was implemented with Pytorch and trained above a NVIDIA RTX3090.  

\subsection{Dataset}
\subsubsection{The KITTI Dataset}
KITTI dataset is a dataset with data collections of car driving. There are 10 sequences including camera, IMU data, and ground-truth pose.  The frequency of IMU data in the synchronized sequence is only 10 Hz. Instead, we use the inertial data from unsynchronized data folder, in which IMU data is 100 Hz, and synchronize data by hand. We selected Sequence 00-08 (03 is not available) as the training set to train our model and used Sequence 09 and 10 as test set. The frequency of images is 10 Hz and the ground-truth pose is provided at the same timestep as images.

\subsubsection{Our Robot Dataset}
\begin{figure}[h!]
    \centering
    \includegraphics[scale=.4]{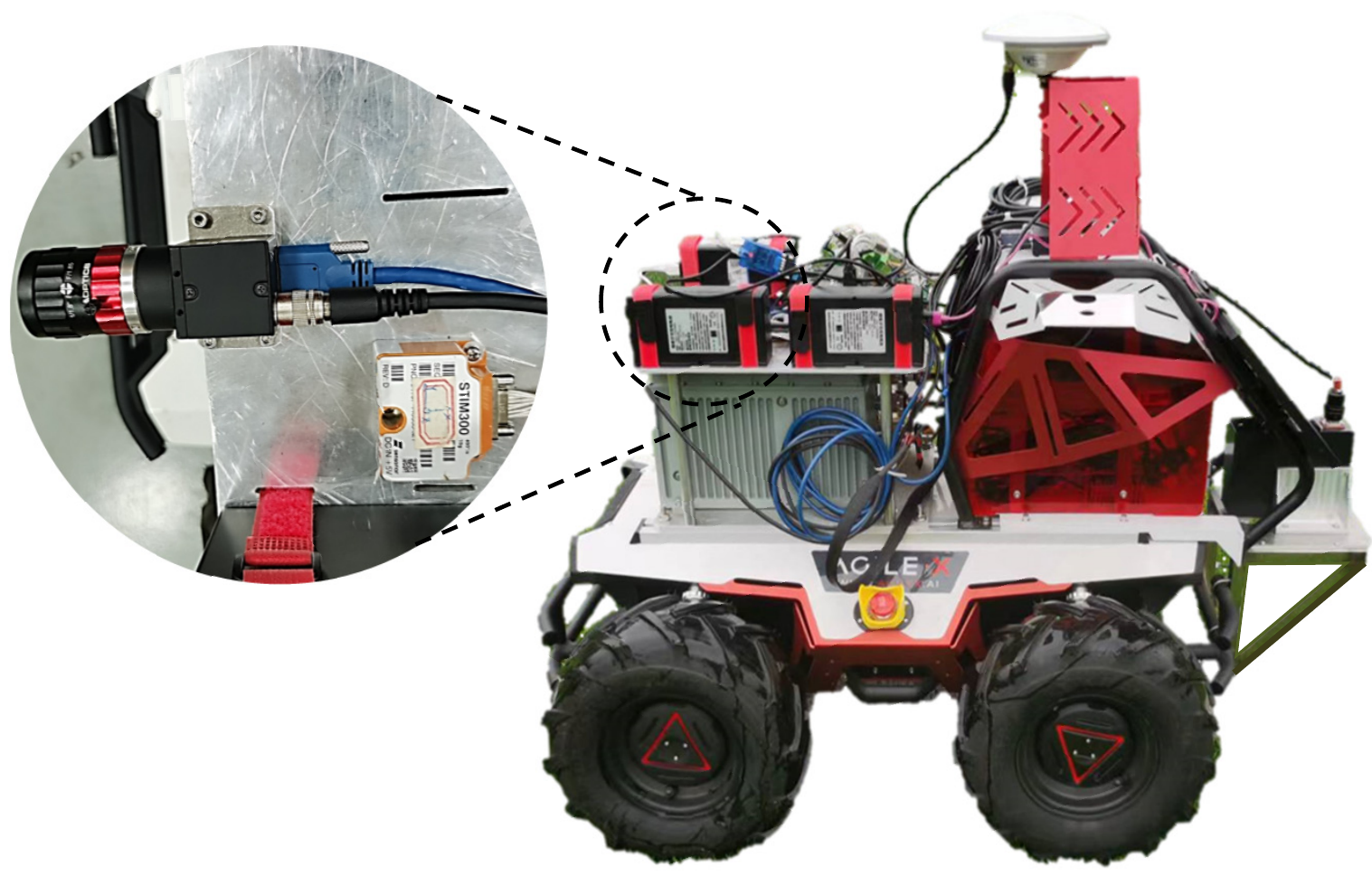}
    \caption{We evaluated our proposed learning based VIO model with our own ground-vehicle platform equipped with RGB camera, IMU, and GNSS/IMU system.}
    \label{fig:equipment}
    \end{figure}

\begin{figure}[!htbp]
    \centering
    \subfigure[Scene1]{
        \label{fig:scene1} 
        \includegraphics[width=3.8cm]{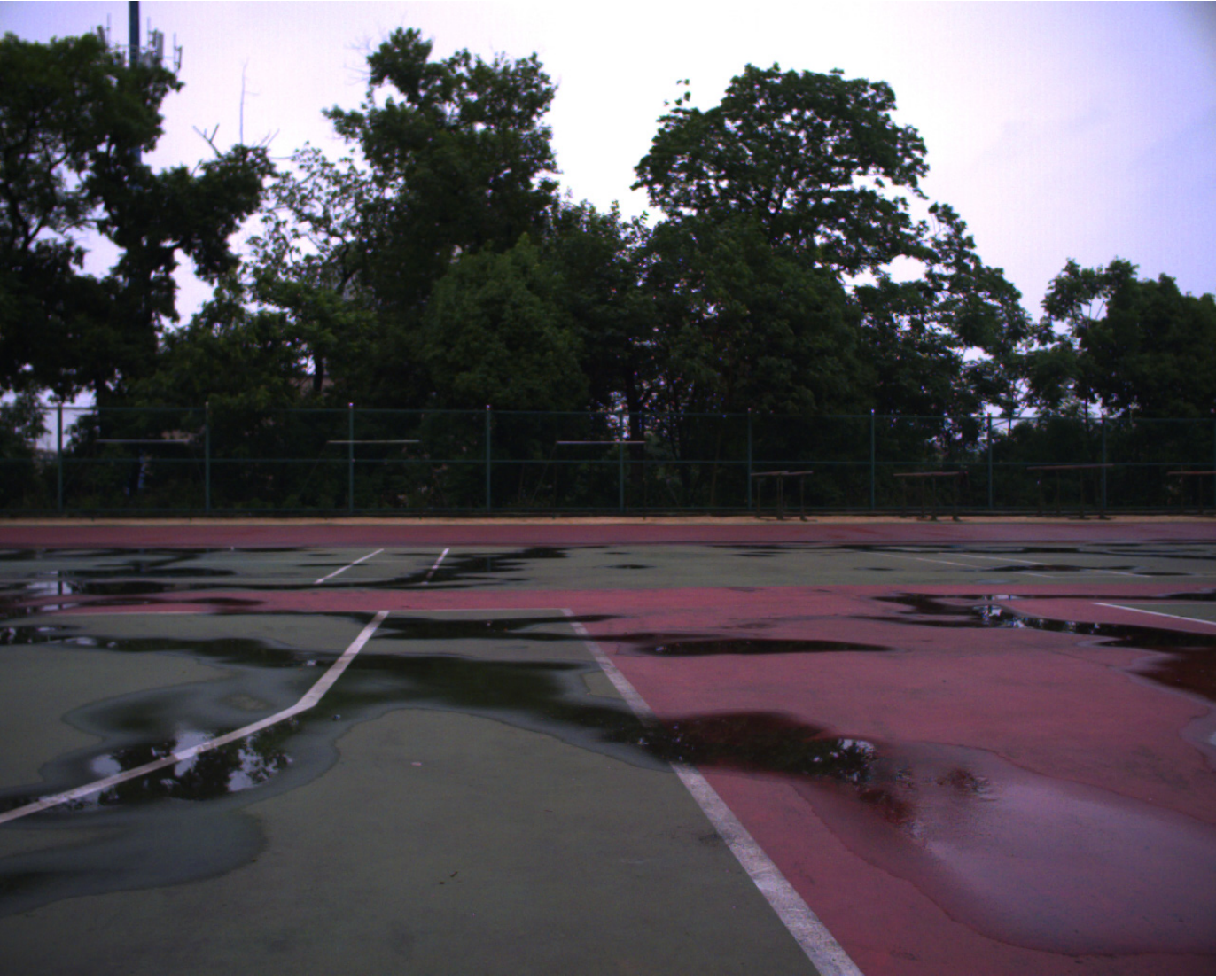}}
    \subfigure[Scene2]{
        \label{fig:scene2} 
        \includegraphics[height=3.06cm,width=3.8cm]{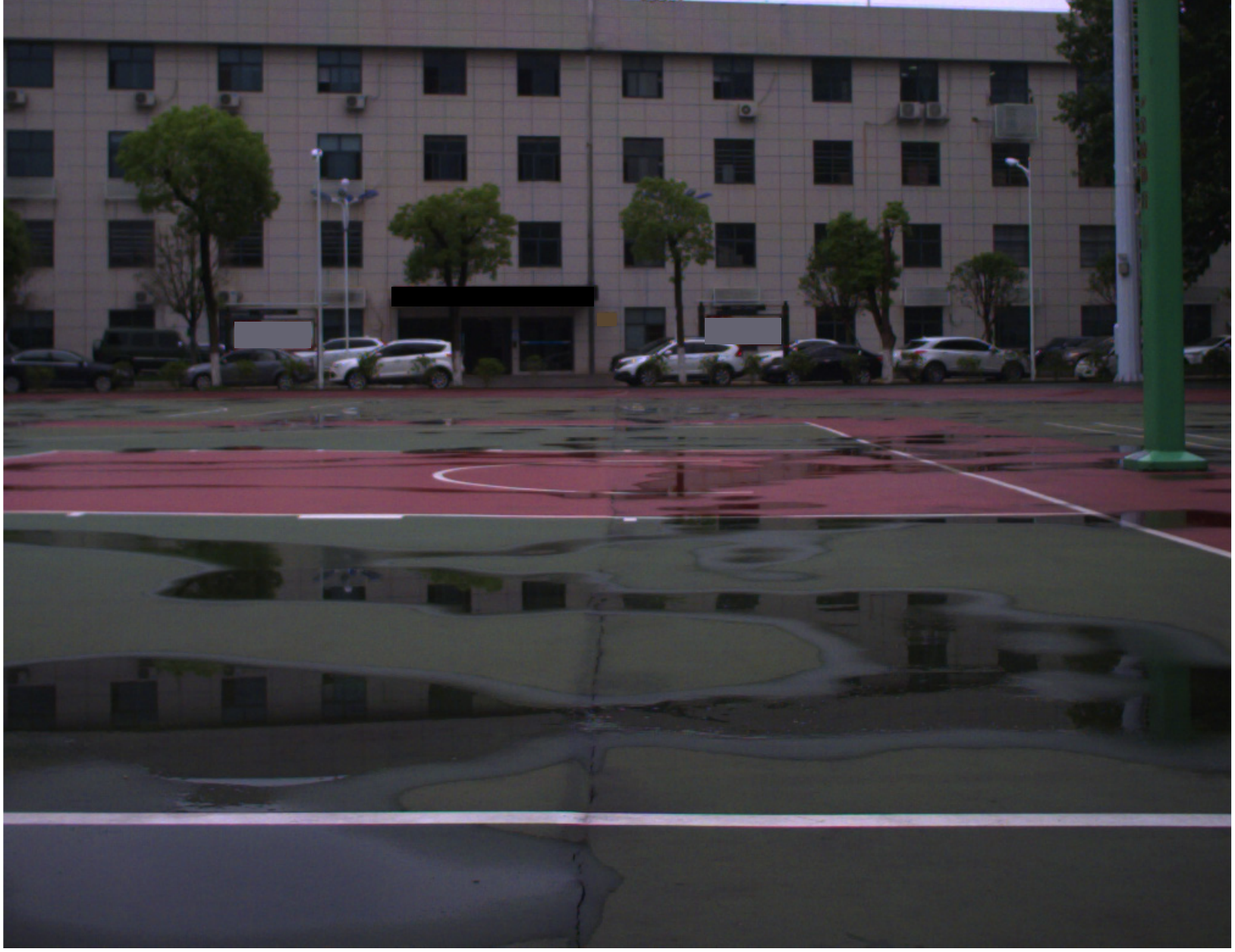}}
        \vfill
        \subfigure[Scene3]{
        \label{fig:scene3} 
        \includegraphics[width=3.8cm]{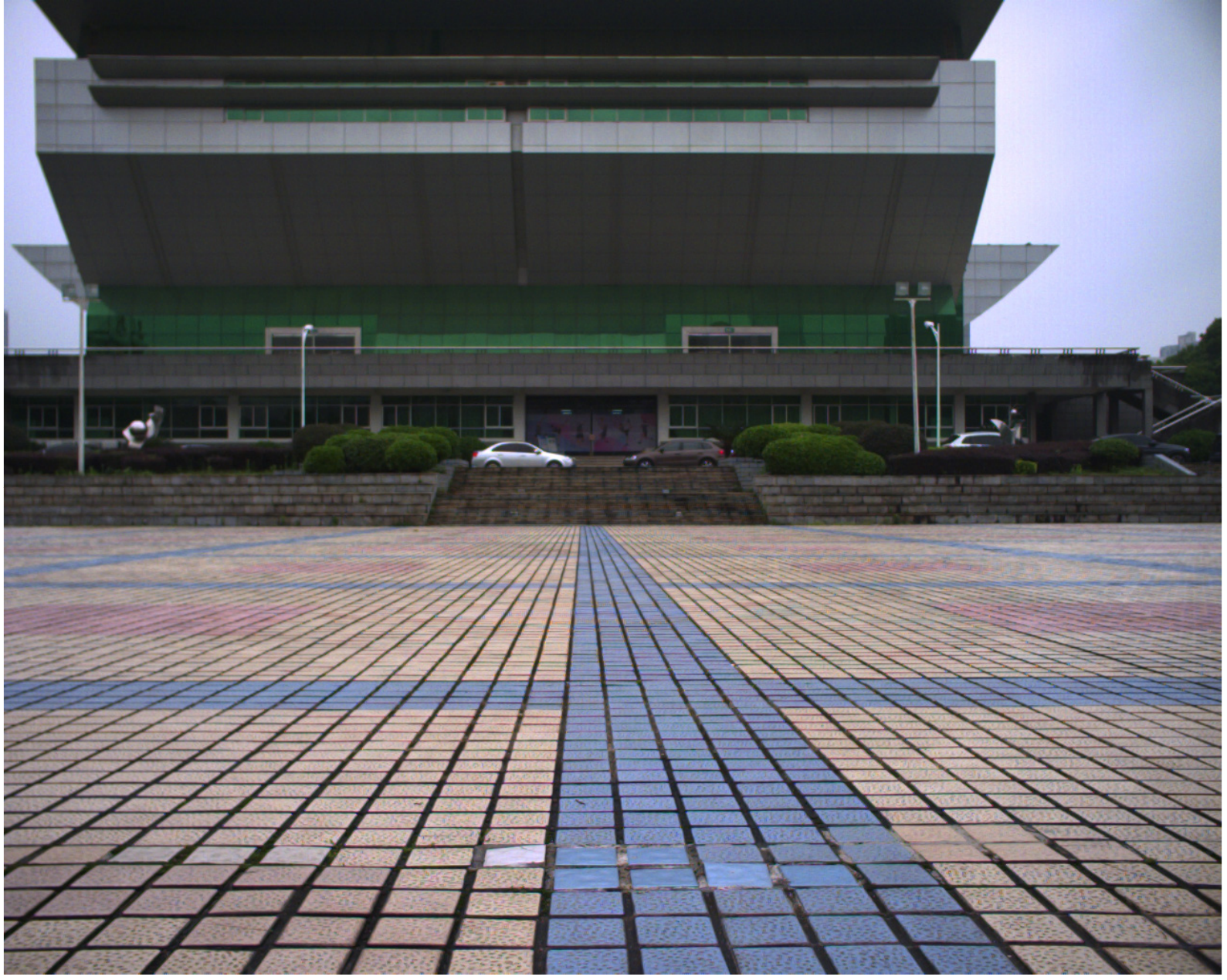}}
    \subfigure[Scene4]{
        \label{fig:scene4} 
        \includegraphics[width=3.8cm]{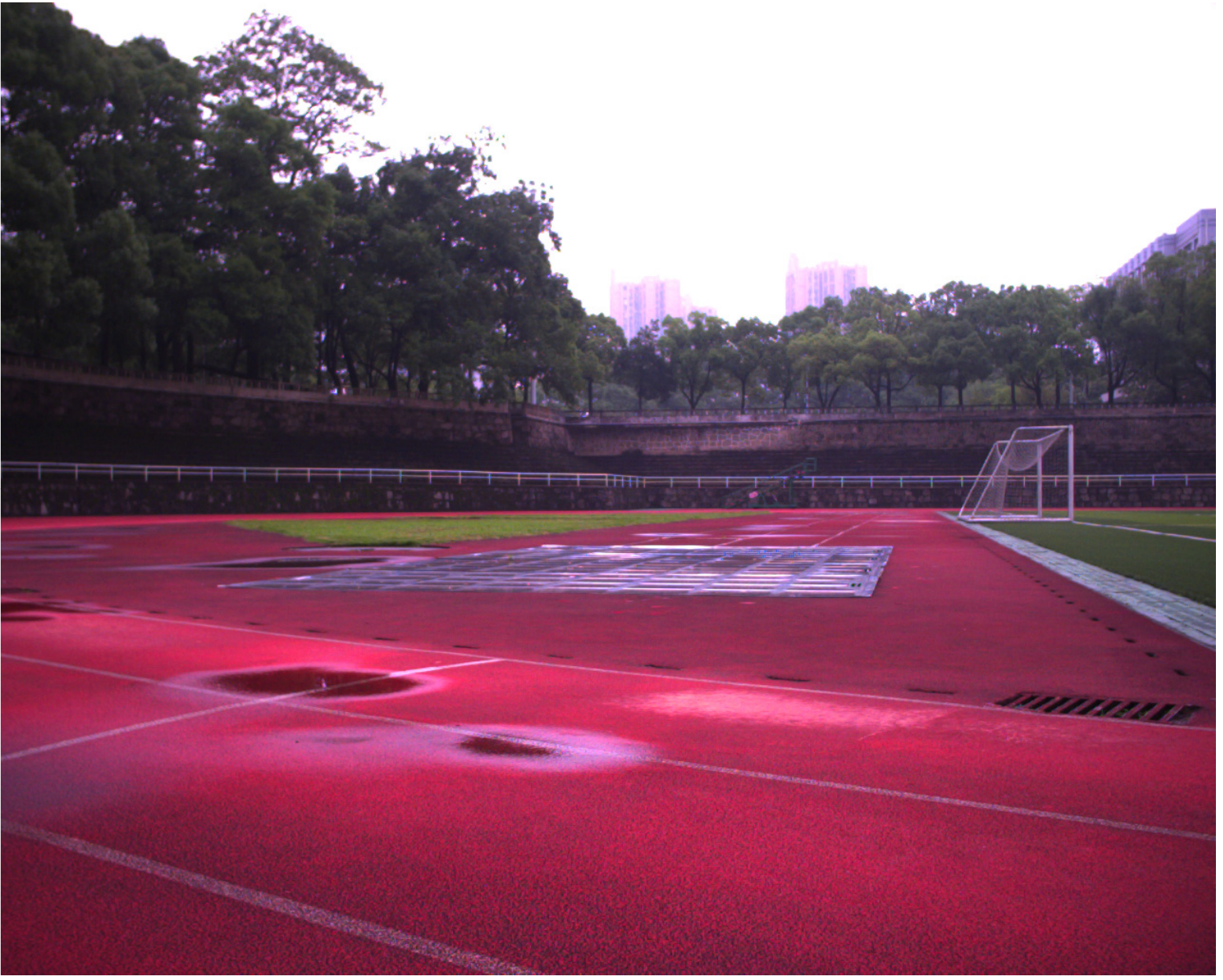}}
    \caption{Our experiment was conducted on overcast days with relatively poor lighting conditions and a large amount of water on wet ground that will cause light reflections. These scenes are difficult for visual feature extraction.}
    \label{fig:scene}
    \end{figure}

We used a ground vehicle platform equipped with a monocular RGB camera and a MEMS IMU, as shown in Fig. \ref{fig:equipment}, to collect images in 10 Hz and inertial data in 100 Hz.
Meanwhile, to capture ground-truth pose, an integrated navigation system with high-precision fiber IMU and GNSS was adopted. The pose accuracy provided by INS/GNSS system is around 0.1 m. The experiment was conducted on overcast days with relatively poor lighting conditions and a large amount of water on wet ground that will cause light reflections, as shown in Fig \ref{fig:scene}.
This scene is difficult for traditional VIOs to extract and match visual features. Our collected dataset includes 7 sequences (00-06). We used the two of them as test set and the rest as the training and validation set.

\subsection{Positioning Results on the KITTI Dataset}
\begin{table*}[h!]
    \centering
    \caption{Positioning Results on the KITTI dataset.}
    \renewcommand{\arraystretch}{1.25}
    \begin{tabular}{ccccccccccc}
    \toprule
    \multirow{2}{*}{Seq.}&  \multicolumn{2}{c}{\added{OKVIS}}   & \multicolumn{2}{c}{VINS-Mono}   &\multicolumn{2}{c}{\added{Zhan et al.}}& \multicolumn{2}{c}{VINet}       & \multicolumn{2}{c}{EMA-VIO (Ours)} \\ \cline{2-11} 
    & \added{$t_\text{rel}$(\%)} &\added{$r_\text{rel}$($^\circ$)} & $t_\text{rel}$(\%) & $r_\text{rel}$($^\circ$) &\added{$t_\text{rel}$(\%)} & \added{$r_\text{rel}$($^\circ$)} &  $t_\text{rel}$(\%) & $r_\text{rel}$($^\circ$) & $t_\text{rel}$(\%)   & $r_\text{rel}$($^\circ$)  \\ \hline
    09 &   \added{9.77}    &     \added{2.97}    &   41.47    &     2.41  &  \added{11.92}    &     \added{3.60}      &      11.83     &     3.00  &    \textbf{8.68} &  \textbf{1.54}  \\
    10 &  \added{17.30}   &     \added{2.82}       &  20.35   &     2.73       & \added{12.62}   &     \added{3.43}&    8.60       &    4.39   &    \textbf{7.46} &  \textbf{2.26}  \\
    average  & \added{13.51} &     \added[id=A1]{2.90}       &30.91 &     2.57   & \added{12.27} &     \added{3.52}    &    10.22       &    3.69   &    \textbf{8.07} &  \textbf{1.90}  \\ \bottomrule
    \multicolumn{11}{p{320pt}}{*$t_\text{rel}$ and $r_\text{rel}$ indicate the averaged translation and rotation drift of all subsequences with length of (100, ..., 800 m).}
    \end{tabular}
    \label{tb: kitti}
    \end{table*}

\begin{figure}[!htbp]
    \centering
    \subfigure[Sequence 09]{
        \label{fig:kitti:09} 
        \includegraphics[width=6cm]{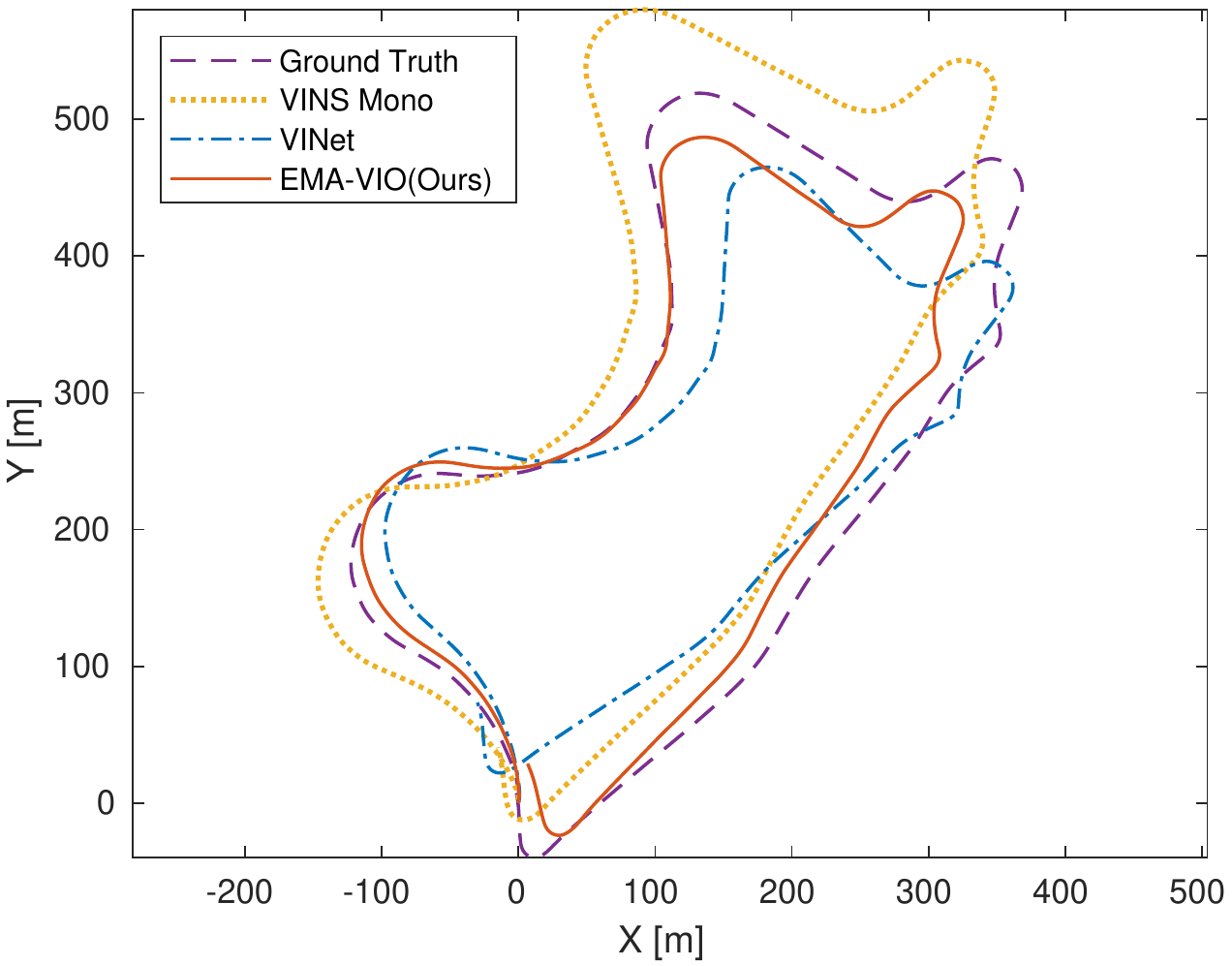}}
    \subfigure[Sequence 10]{
        \label{fig:kitti:10} 
        \includegraphics[width=6cm]{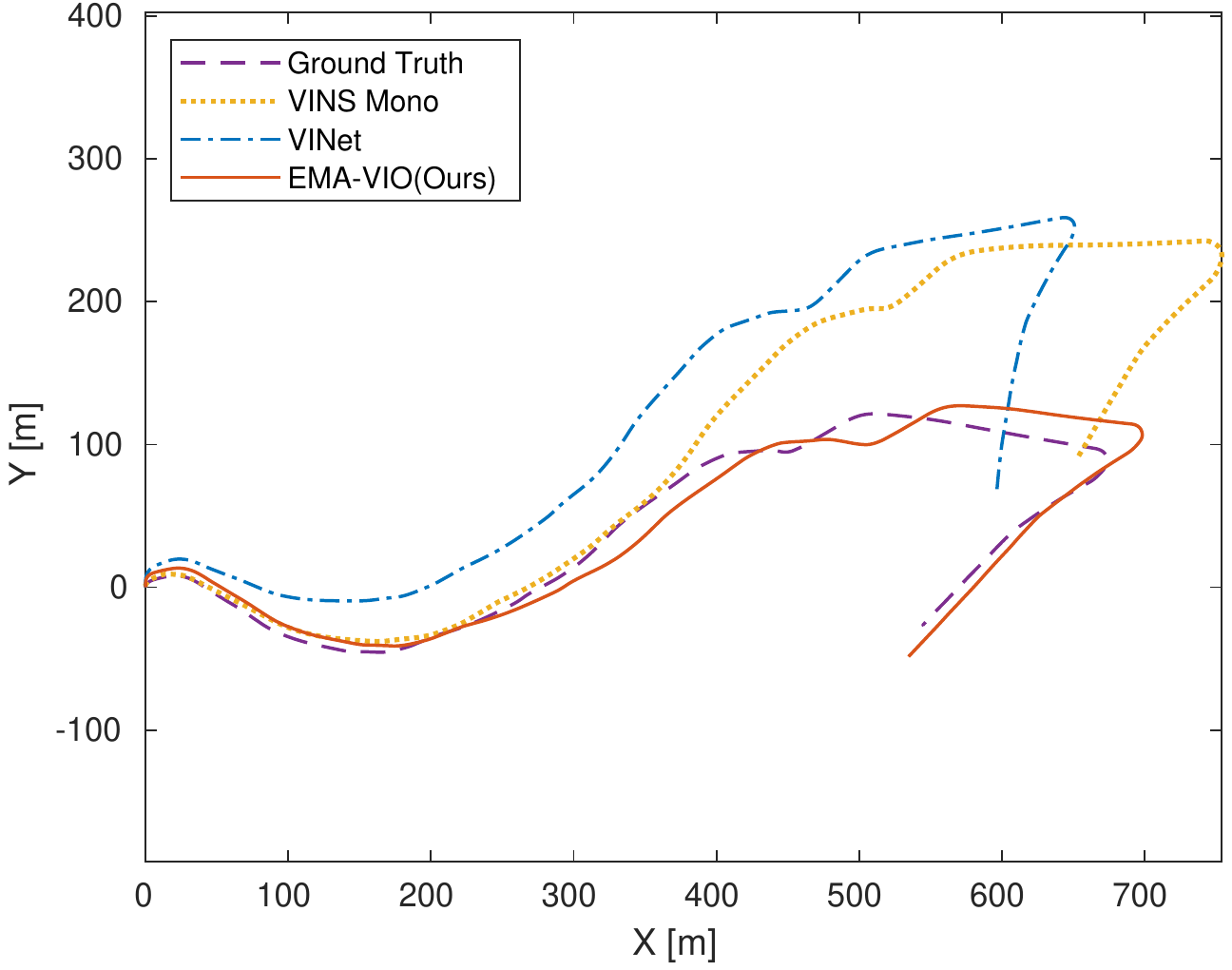}}
    \caption{The positioning results on the KITTI dataset. Our proposed learning based VIO model outperforms both traditional VINS-mono and learning based VINet.}
    \label{fig: kitti}
    \end{figure}

\added{We first evaluated our proposed EMA-VIO on the KITTI dataset, and compare the pose results of our EMA-VIO with two traditional VIO methods (i.e., VINS-Mono\cite{qin_vins-mono_2018}, OKVIS\cite{leutenegger2013keyframe}) and two learning based pose estimation methods (i.e.  VINet\cite{clark_vinet_2017} and \cite{Zhan2018}).}
VINet extracts visual features using FlowNet, but uses LSTM to extract inertial features, and fuses visual and inertial features with LSTM. From Figure \ref{fig: kitti}, clearly, the trajectories of our proposed EMA-VIO on Sequence 09 and 10 of KITTI dataset are closer to the ground-truth trajectories of these two sequences.

Furthermore, as demonstrated in Table \ref{tb: kitti}, the official evaluation metric of KITTI dataset is adopted to compare the quantitative results of three VIO models, which is the average translational RMSE drift(\%) and the average rotational RMSE drift($^\circ$/100m). 
It can be found that our method outperforms VINS-mono and VINet in terms of both translation and rotation. 
Compared with VINet, our proposed learning based VIO significantly reduces the translation drift around 20\%, and the rotation drift by 48\%.
Both VINet and our EMA-VIO show advantages over the VINS-Mono, as cameras and IMU are not strictly calibrated and time-synchronized on the KITTI dataset, and this causes troubles to the traditional hand-designed VIO system.
\added{It can be seen that the introduced external memory attention effectively integrates visual and inertial features, that contributes to build potential relationship among different features and adaptively adjust the feature weights at different place according to the training data. Moreover, the introduced memory unit improves the ability of the attention module to deal with the feature with time sequence information, so that it can be used to replace the decoder based on LSTM in the previous learning-based visual-inertial odometry.}

\subsection{Positioning Results on Our Robot Dataset}
\begin{figure}[h!]
    \centering
    \subfigure[Test Sequence 1]{
        \label{fig:our:03} 
        \includegraphics[width=6cm]{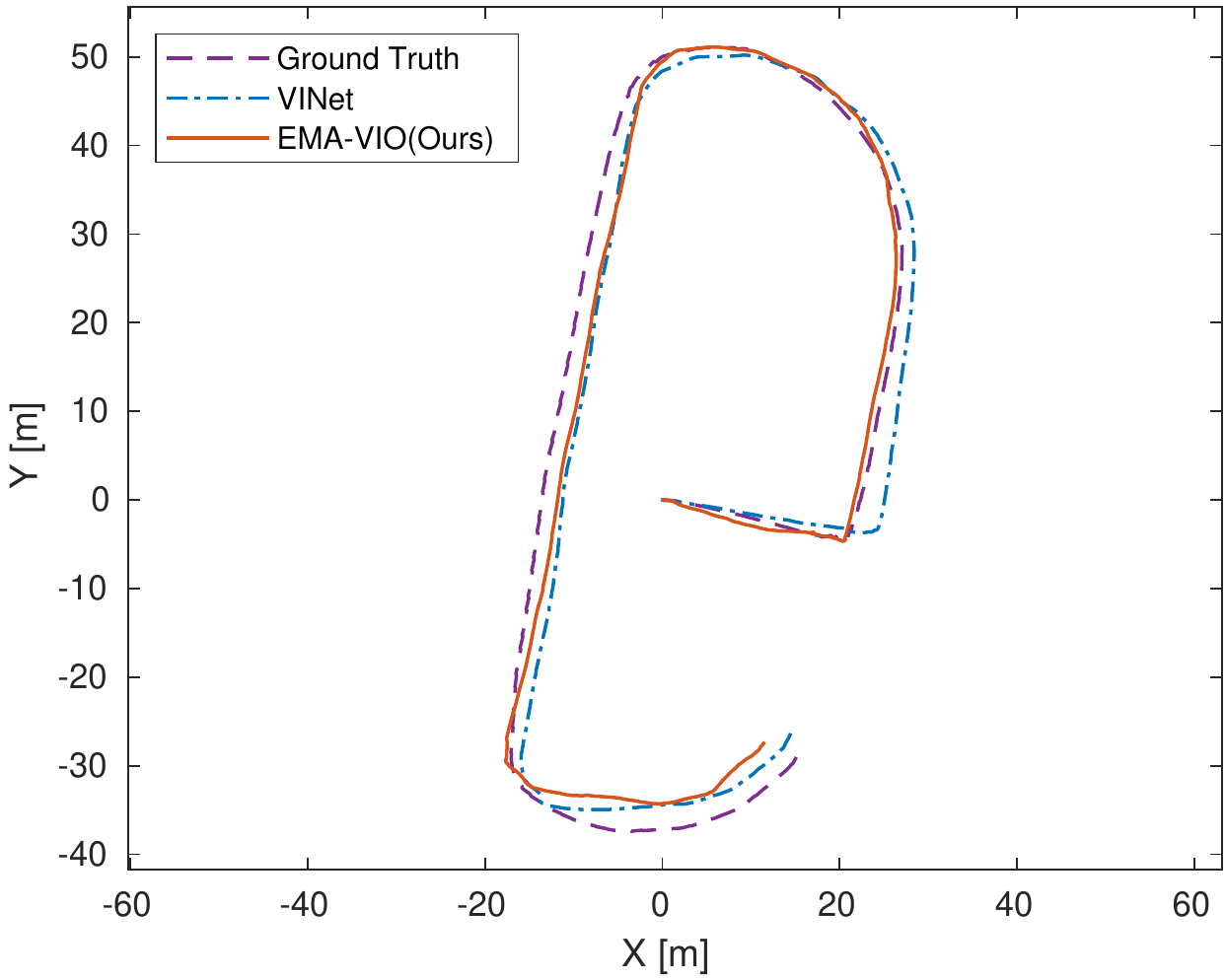}}
    \subfigure[Test Sequence 2]{
        \label{fig:our:10} 
        \includegraphics[width=6cm]{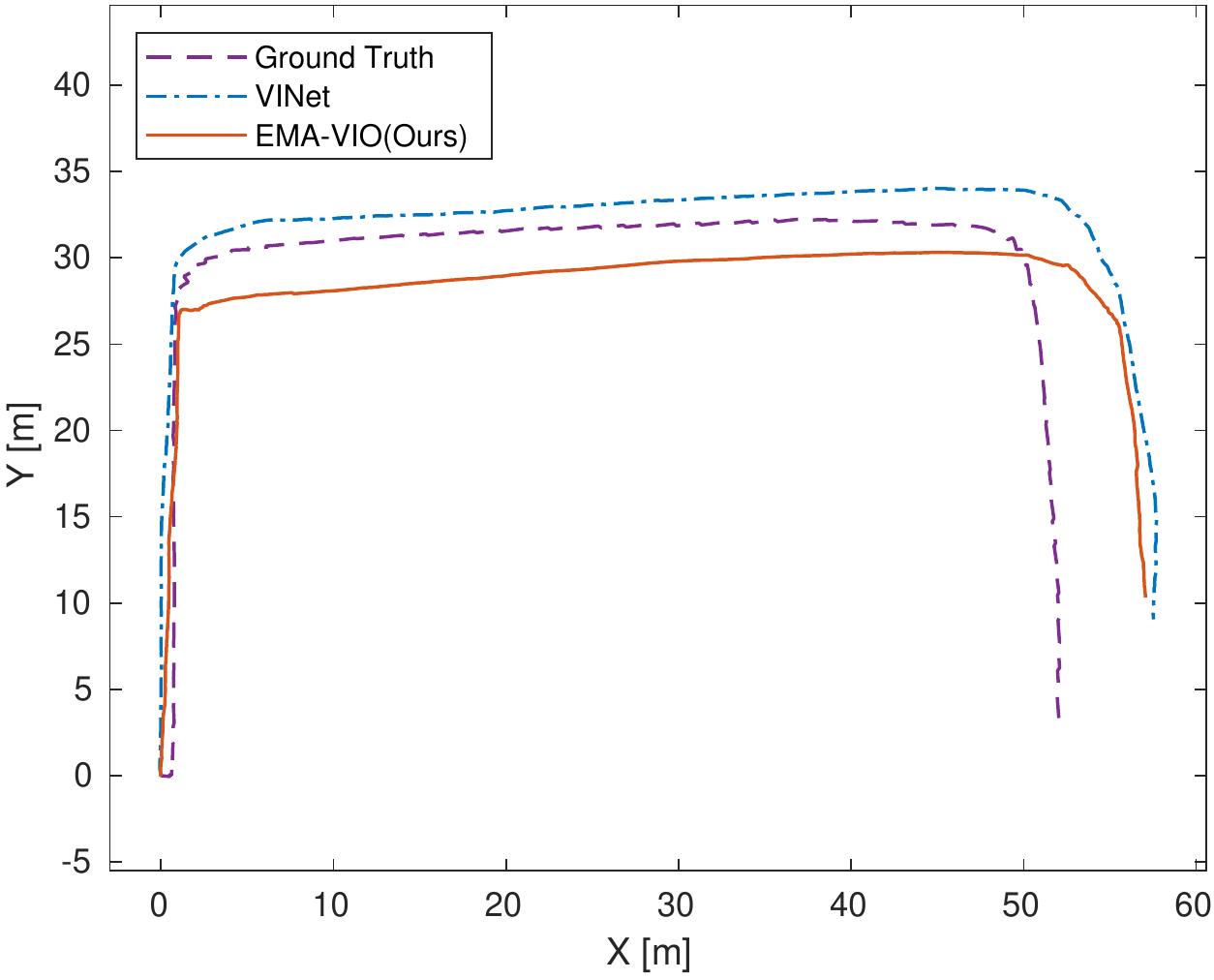}}
    \caption{The positioning results of our proposed EMA-VIO on our robot dataset in challenging scenes, compared with VINet. VINS-mono failed on this dataset as its extracted visual features are not enough to perform algorithm.}
    \label{fig: ourdataset}
    \end{figure}
    
    \begin{figure}[!htbp]
    \centering
    \subfigure[Scene 1]{
        \label{fig:vins03} 
        \includegraphics[width=4cm]{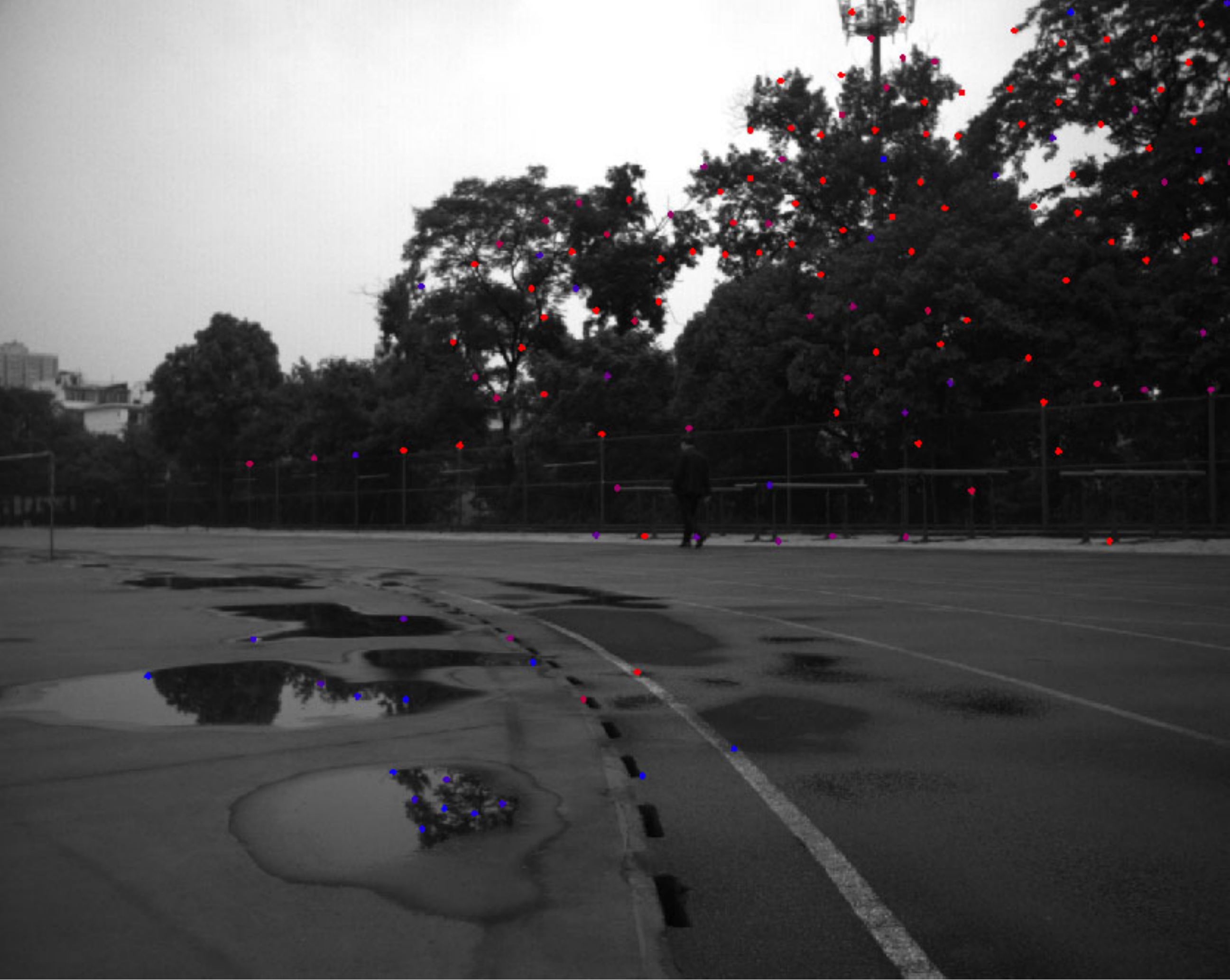}}
    \subfigure[Scene 2]{
        \label{fig:vins10} 
        \includegraphics[width=4cm]{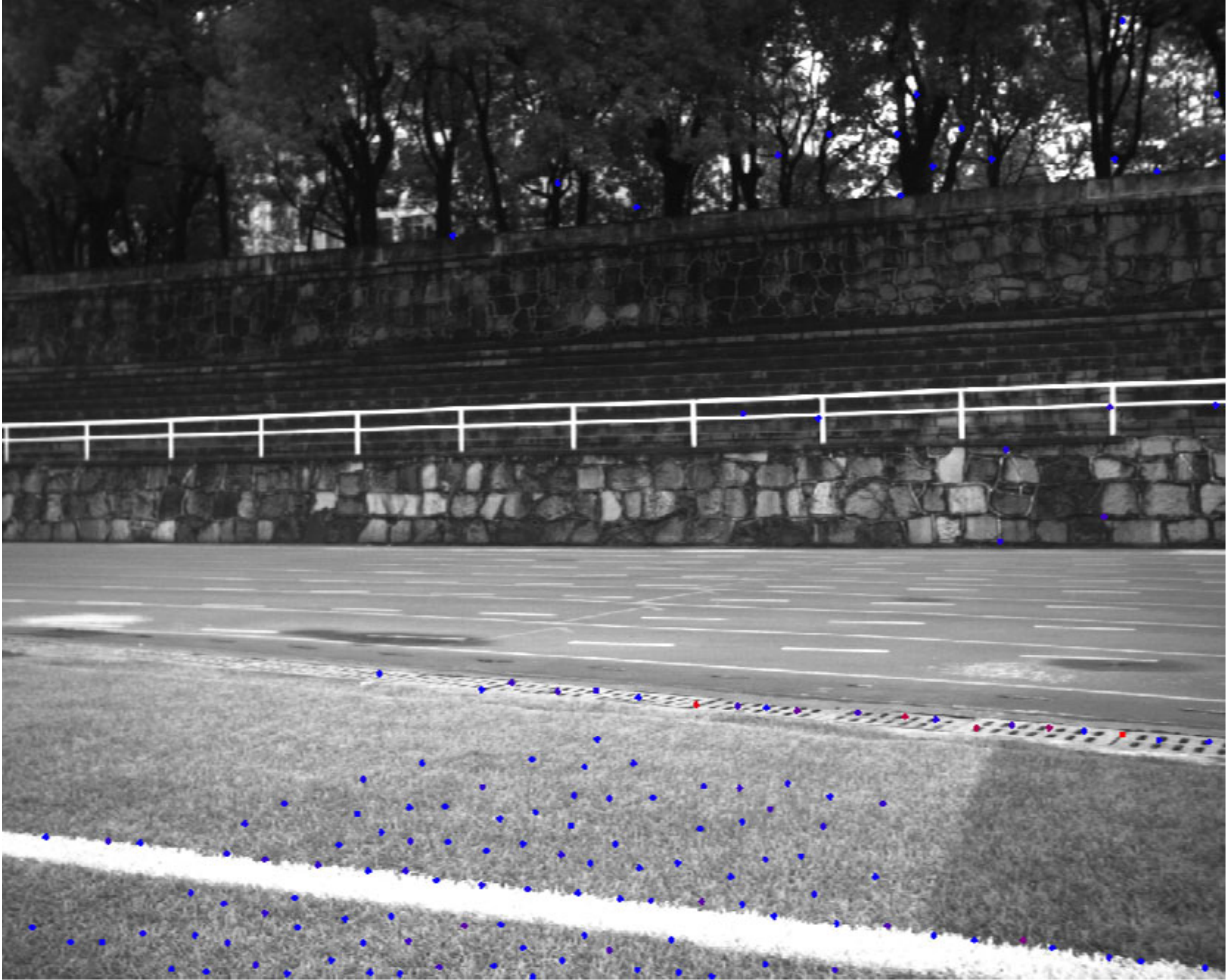}}
    \caption{The extracted features of VINS-Mono algorithm}
    \label{fig:ourvins}
    \vspace{-0.3cm}
    \end{figure}

Then, an outdoor robot navigation experiment was conducted to collect data for the evaluation of our proposed VIO model. As 
the height of our groundtruth pose from GNSS/IMU system is not accurate enough, we mainly evaluate positioning results on the plane. Figure \ref{fig: ourdataset} shows the trajectories of our proposed EMA-VIO and a baseline, i.e., VINet. Both of them can perform well in these challenging scenes. We find that VINS-Mono fails to extract and track feature points in the scenes of our dataset, as poor lighting condition of the overcast day and the water on the ground with light reflections cause troubles to the hand-designed feature detectors. As shown in Figure \ref{fig:ourvins}, the extracted visual features of VINS are not enough to perform positioning algorithm. This indicates that deep neural networks are more powerful to extract features for localization.
The quantitative results are shown in Table \ref{tab:ourdata}.
The evaluation metric we used here is the horizontal position error(HPE) according to Eq. \ref{eval}.
\begin{equation}\label{eval}
    \text{Error} = \sqrt{\frac{1}{n} (\mathbf{p}_i - \hat{\mathbf{p}_i})^2} 
\end{equation}
where \added{n is the length of the trajectory sequence,} \textbf{$\mathbf{p}_i$} denotes the planar localization result $(x_i,y_i)$ of the estimated value of the algorithm and \textbf{$\hat{\mathbf{p}_i}$} denotes the planar localization value $(\hat{x_i},\hat{y_i})$ of the ground truth.
Clearly, our EMA-VIO outperforms VINet in positioning results.
This indicates the effectiveness of our introduced external memory attention and geometric constraints. 
\added{Our method shows good robustness in the challenging navigation scene. 
Though traditional VIO method performs well in the scenes with rich texture and demonstrates good localization results, the hand-crafted features are hard to be applied in challenging navigation scenes (e.g. with sparse features and poor light conditions), and thus traditional VIO approaches usually perform not well due to the lack of effective feature tracking. The deep learning-based VIO approach continues to show good performance in the challenging experimental scenerios due to the learned features.
}
    
\begin{table}[htbp]
    \centering
    \caption{Results on the testing sequences of our robot dataset in challenging environments.}
    \begin{tabular}{cccc}
    \toprule
            & VINS-mono & VINet & EMA-VIO (ours) \\ \midrule
    Test 01   & Fail     &    2.9080 m    &    \textbf{2.3401 m}         \\
    Test 02  & Fail      &   4.3956 m    &    \textbf{4.2437 m}         \\
    average  & Fail  &  3.6518 m    &    \textbf{3.2919 m}           \\ \bottomrule
    \end{tabular}
    \label{tab:ourdata}
    \end{table}
    \vspace{-0.3cm}
    
        \begin{table}[h!]
        \centering
        \caption{Ablation study results of our model}
        \setlength{\tabcolsep}{0.9mm}{
        \renewcommand{\arraystretch}{1.25}
        \begin{tabular}{cccccccccc}
        \toprule
         \multirow{2}{*}{VIO} & \multirow{2}{*}{EA} & \multirow{2}{*}{WaveNet} & \multirow{2}{*}{MC} & \multicolumn{2}{c}{Seq09} & \multicolumn{2}{c}{Seq10} & \multicolumn{2}{c}{Avg} \\ \cline{5-10} 
            &    &    &    & $t_\text{rel}(\%)$ & $r_\text{rel}$($^\circ$) & $t_\text{rel}(\%)$ & $r_\text{rel}$($^\circ$) & $t_\text{rel}(\%)$ & $r_\text{rel}$($^\circ$)  \\ \hline
         $\checkmark$ &             &        &           & 11.83 & 3.00 & 8.60 & 4.39 & 10.22 & 3.69  \\
         $\checkmark$ & $\checkmark$ &       &           & 10.23 & 1.33 & 9.44 & 2.06 & 9.83 & 1.69  \\
         $\checkmark$ &   & $\checkmark$     &           & 9.38 & 1.94 & 11.07 & 4.29 & 10.22 & 3.11 \\
         $\checkmark$ & $\checkmark$ &  & $\checkmark$   & 9.34 & \textbf{1.09} & 7.94 & \textbf{1.61} & 8.64 & \textbf{1.35}   \\ 
         $\checkmark$ & $\checkmark$ & $\checkmark$ & $\checkmark$ & \textbf{8.68} & 1.54 & \textbf{7.46} & 2.26 & \textbf{8.07} & 1.90 \\ 
         \bottomrule
         \multicolumn{10}{p{240pt}}{*$t_\text{rel}$ and $r_\text{rel}$ indicate the averaged translation and rotation drift of all subsequences with length of (100, ..., 800 m).}
        \end{tabular}}
        \label{ablation}
        \end{table}
        \vspace{-0.3cm}
    
\subsection{Ablation Study}

    In order to validate the performance improvement of each module in our proposed approach, we also conducted an ablation study of our EMA-VIO model on the KITTI dataset. The results are shown in the Tab. \ref{ablation}.
    VIO indicates the baseline learning based VIO with LSTM for multimodal fusion and inertial data processing, while EA means replacing LSTM with external memory attention for feature fusion, WaveNet means replacing LSTM with WaveNet for inertial data processing, and MC means adding multistate geometric constraints. \added{Meanwhile, due to further utilization of the transformation relationship between successive image frames, the convergence speed of the model is increased by introducing the multi-state constraint in the training process. 
    }

    It can be seen that our introduced external attention, WaveNet and multistate geometric constraints further improve the positioning performance base model. Though the WaveNet degrades the rotation slightly,  the introduction of WaveNet still improves positioning results, and improves the model efficiency that will be discussed as below, so that we view it as a tradeoff to use WaveNet. 
    
 \begin{figure}[!htbp]
    \centering
    \subfigure[Parameters Number]{
        \label{fig:parameters} 
        \includegraphics[width=4cm]{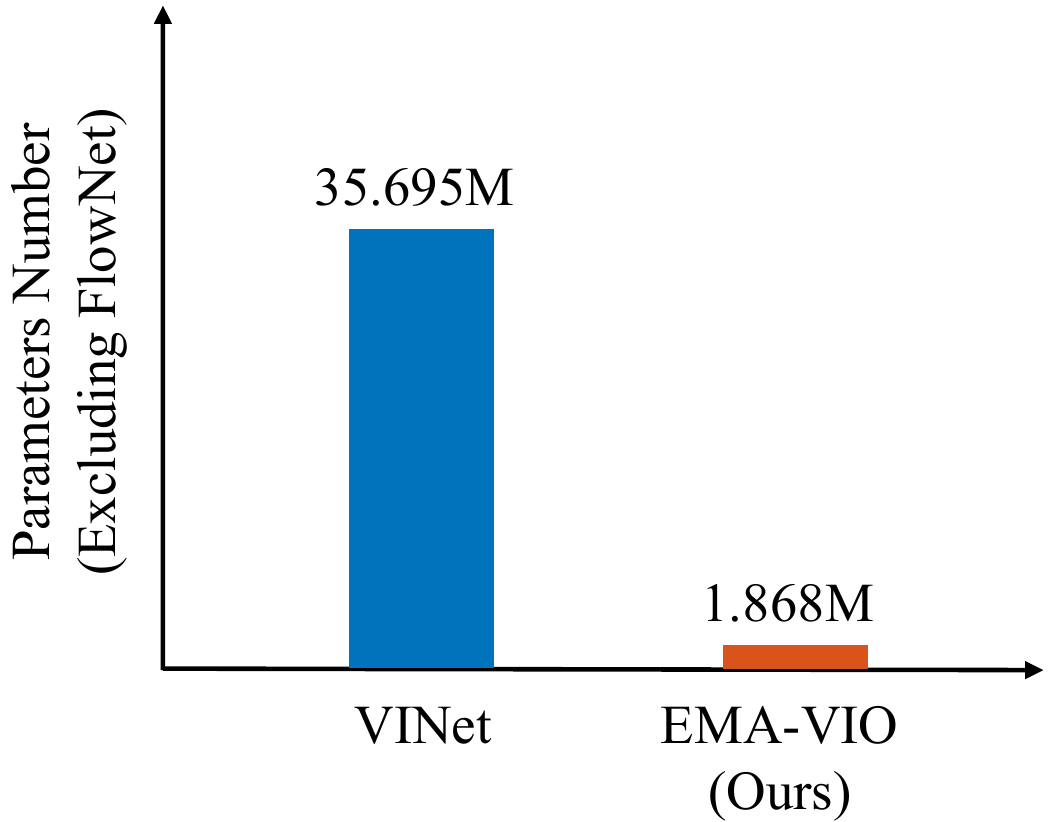}}
    \subfigure[MACs]{
        \label{fig:macs} 
        \includegraphics[width=4cm]{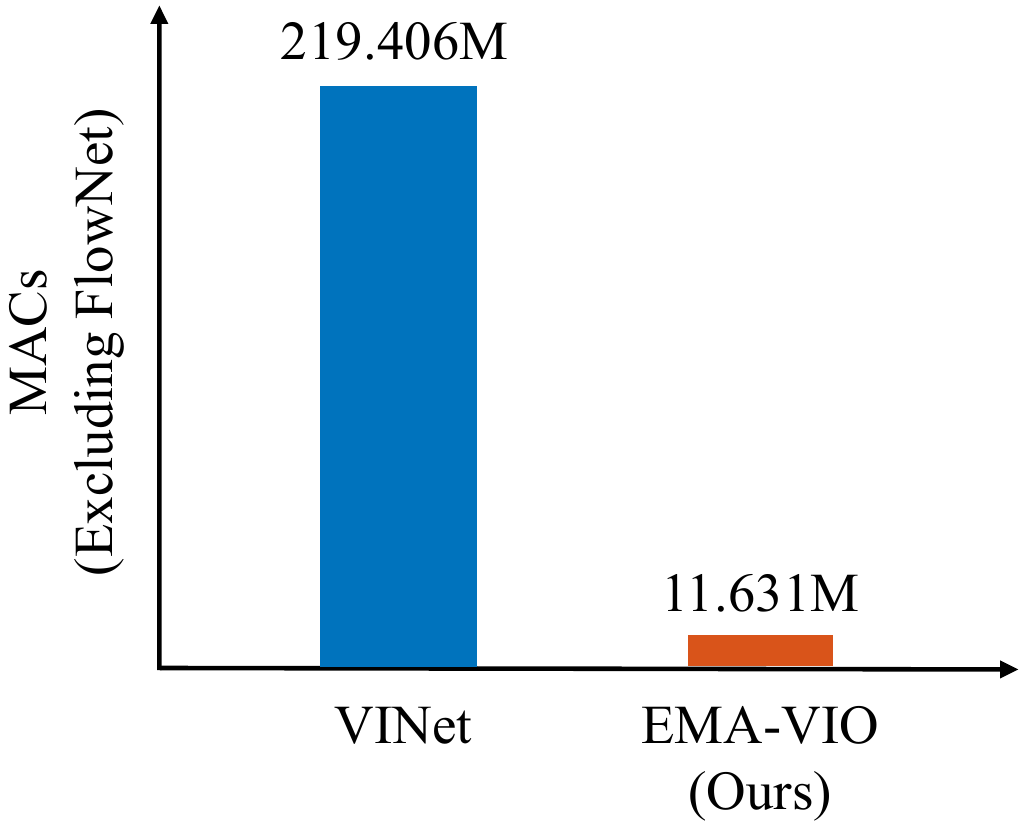}}
    \caption{The comparison of our EMA-VIO model and VINet in terms of model parameters number and MACs.}
    \label{fig:efficiency}
    \end{figure}
    \vspace{-0.3cm}

\subsection{Efficiency Analysis}

Finally, we come to analyse the efficiency of our proposed EMA-VIO. Two important terms are selected to compare with VINet. One is the number of parameters that represents the memory consumption, and the other one is multiply-accumulate operations (MACs) that represents the computation requirement.
As FlowNet in both our EMA-VIO and VINet occupies large resources of memory and computation, we excludes FlowNet in our calculation, considering this visual extraction module can be replaced with new technique in the future, that is not a main point in this work.
Our EMA-VIO method shows great advantages not only in parameters number but also in MACs, compared with VINet as shown in Fig. \ref{fig:efficiency}. Our model reduces the parameters number from 35.695M to 1.868M, and reduces the MACs from 219.406M to 11.631M. This is due to the fact that we replace LSTM with external memory attention for feature fusion, and replace LSTM with WaveNet for inertial data processing.

\section{Conclusion}
In this work, we proposed a novel learning based VIO framework with external memory attention for visual and inertial sensor fusion. Additionally, multi-state geometric constraint is introduced into our framework to further optimize the pose estimation. Real-world experiments on public dataset and our own robot platform show that our proposed EMA-VIO outperforms both traditional and learning based VIO models. Moreover, our model is more efficient and reduces the memory and computational requirement. \added{However, our model still requires to be trained on the GPU, limiting its further applications in the micro navigation systems. Besides, the stability of our proposed model needs to be verified in open environment.
}
\bibliographystyle{IEEEtran}
\bibliography{citationlist}
\end{document}